\documentclass[letterpaper, 10 pt, conference]{ieeeconf}  

\IEEEoverridecommandlockouts                              

\usepackage{graphicx} 
\usepackage{mathptmx} 
\DeclareMathAlphabet{\mathcal}{OMS}{cmsy}{m}{n}
\DeclareSymbolFont{largesymbols}{OMX}{cmex}{m}{n}
\usepackage{amsmath} 
\usepackage{amssymb}  
\usepackage{amsfonts}
\usepackage{subfig}
\usepackage{cite} 
\usepackage{booktabs}
\makeatletter
\let\NAT@parse\undefined
\makeatother
\usepackage{hyperref}

\title{\LARGE \bf
EAROL: Environmental Augmented Perception-Aware Planning and Robust Odometry via Downward-Mounted Tilted LiDAR
}

\author{Xinkai Liang$^{1}$, Yigu Ge$^{1}$, Yangxi Shi$^{1}$, Haoyu Yang$^{1}$, Xu Cao$^{1}$ and Hao Fang$^{1,2}$
\thanks{*This work was supported in part by the National Nature Science Foundation of China (NSFC) under Grant (No.62133002) and the Fundamental Research Funds for the Central Universities (No.2024CX06098) separately.}
\thanks{$^{1}$All authors are with School of Automation, Beijing Institute of Technology. Xinkai Liang (yudubai@bit.edu.cn), Yangxi Shi (yangxi.shi@bit.edu.cn)}%
\thanks{$^{2}$The corresponding author: Hao Fang (fangh@bit.edu.cn)}
\thanks{$^{3}$\url{https://github.com/FLAG-BIT/EAROL/}}
}

\begin{document}

\maketitle
\thispagestyle{empty}
\pagestyle{empty}

\begin{abstract}

To address the challenges of localization drift and perception-planning coupling in unmanned aerial vehicles (UAVs) operating in open-top scenarios (e.g., collapsed buildings, roofless mazes), this paper proposes EAROL, a novel framework with a downward-mounted tilted LiDAR configuration (20° inclination), integrating a LiDAR-Inertial Odometry (LIO) system and a hierarchical trajectory-yaw optimization algorithm. The hardware innovation enables constraint enhancement via dense ground point cloud acquisition and forward environmental awareness for dynamic obstacle detection. A tightly-coupled LIO system, empowered by an Iterative Error-State Kalman Filter (IESKF) with dynamic motion compensation, achieves high level 6-DoF localization accuracy in feature-sparse environments. The planner, augmented by environment, balancing environmental exploration, target tracking precision, and energy efficiency. Physical experiments demonstrate 81\% tracking error reduction, 22\% improvement in perceptual coverage, and near-zero vertical drift across indoor maze and 60-meter-scale outdoor scenarios. This work proposes a hardware-algorithm co-design paradigm, offering a robust solution for UAV autonomy in post-disaster search and rescue missions. We will release our software and hardware as an open-source package$^{3}$ for the community. Video: \url{https://youtu.be/7av2ueLSiYw}.
\end{abstract}


\section{INTRODUCTION}
Unmanned Aerial Vehicles (UAVs) are currently widely used in various fields such as industry, agriculture, rescue operations, and photography\cite{rabta2018drone, mishra2020drone, daud2022applications}. In these applications, SLAM module is crucial, which can provide a dense 3D map for navigating in unfamiliar environment safely and accurate state estimation for acquiring the position and attitude of UAVs. Common sensors are cameras \cite{qin2018vins, campos2021orb} and LiDAR. However, due to the cameras' inability to directly obtain accurate three-dimensional information and its limitation in low-light conditions, LiDAR is increasingly becoming a powerful sensor for these tasks\cite{li2021towards, wang2021lightweight}. Nevertheless, LiDAR SLAM algorithms face challenges in extreme scenarios, such as overly open environment or rubble rescue situations where there are no distinct features above. In these degenerate scenes, a common-mounted LiDAR lacks constraints, leading to divergence and drift in state estimation. To address these issues, the technological evolution is shifting from single-sensor optimization to tightly-coupled multi-sensor fusion (e.g., vision-LiDAR-IMU\cite{zhang2017vloam}) and hierarchical feature extraction (e.g., ground/intensity/reflectivity). These advancements are further enhanced by online calibration and dynamic weight allocation to improve system adaptability. However, current approaches predominantly rely on post-processing optimization, exhibiting a lack of co-design between hardware configurations and algorithmic models. This opens up opportunities for innovation in novel sensor configurations and online adaptive calibration to holistically address drift challenges across diverse operating conditions.

\begin{figure*}[t]
\centering
\includegraphics[width=1.0\textwidth]{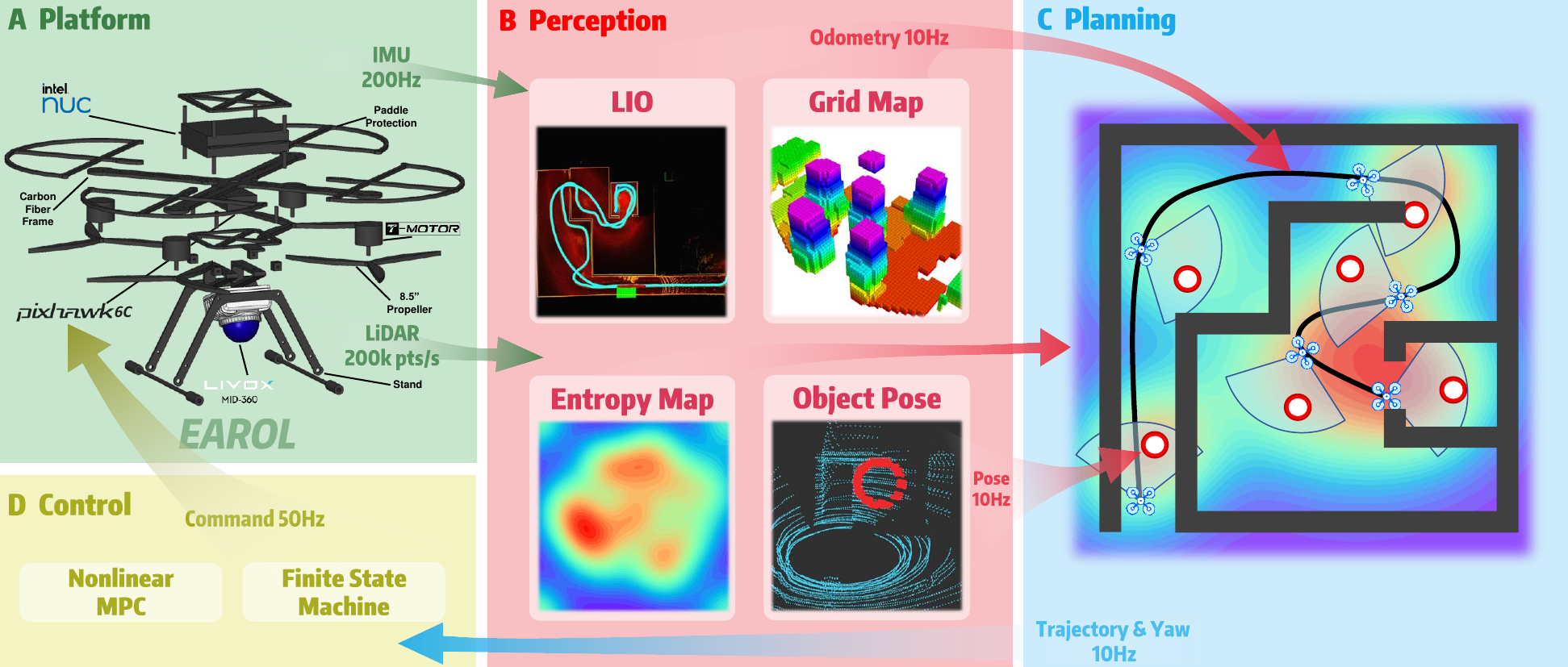}
\caption{System Overview of \textbf{EAROL}. (A) \textbf{Platform:} The exploded view of UAV hardware component. (B) \textbf{Perception:} The LIO module operates at 10Hz pose estimation frequency with tightly-coupled 200Hz IMU data fusion, while simultaneously constructing a multi-layer environmental representation comprising geometric occupancy grid maps and probabilistic entropy maps, along with real-time dynamic object tracking through a robust RANSAC-based segmentation approach. (C) \textbf{Planning:} The trajectory generation module executes spatiotemporal-constrained motion planning within the fused odometric-geometric state space, while simultaneously conducting a computationally efficient yaw angle sequence optimization based on entropy map gradients and real-time RANSAC-tracked object positions. (D) \textbf{Control:} The flight control subsystem processes navigation references through a nonlinear model predictive controller (NMPC) with receding horizon optimization and a behavior-regulated finite state machine (FSM), achieving certified robust control command generation at 50Hz frequency.}       
\label{zongtu}       
\end{figure*}

To address these challenges, we propose a novel downward-tilted LiDAR configuration that strategically reorients the sensor's Field of View (FOV) to enhance 3D perception capabilities. This innovative installation achieves dual functionality: (1) vertical constraint reinforcement through ground surface point cloud acquisition, and (2) forward perception augmentation for dynamic obstacle detection ahead. Building upon this hardware innovation, we develop a robust LIO system featuring dynamic motion compensation and degeneracy-aware processing that automatically adjusts feature weights in low-texture environments.

The directional perception paradigm introduced by this configuration causes the perception area of LiDAR changes from omni-directional to directional, which creates a critical perception-planning coupling challenge: conventional LiDAR-based planners typically generate yaw angles passively aligned with trajectory tangents, neglecting active perception optimization. Inspired by visual perception-aware planning principles, we propose an environmental augmented perception-aware yaw angle planning approach.

\textbf{Our main contributions are as follows:}
\begin{itemize}
\item Proposed $\mathbf{EAROL}$, an $\mathbf{E}$nvironmental $\mathbf{A}$ugmented perception-aware planning and $\mathbf{R}$obust $\mathbf{O}$dometry via Downward-Mounted tilted $\mathbf{L}$iDAR framework.
\item Developed an efficient and robust LIO in overly open environment with dynamic compensation via IESKF.
\item Developed a trajectory and yaw planner driven by environment information entropy and tracking error augmented.
\item Constructed a complete UAV system and conducted physical experiments to validate our proposed solution.
\end{itemize}

\section{RELATED WORK}

\subsection{LiDAR SLAM Drift}
Present works on LiDAR SLAM almost origin from LOAM\cite{zhang2014loam}, which achieves 6-DoF localization through edge and planar feature extraction, but their reliance on structural scene features may cause drift. SLAM drift may occur under various scenarios, such as insufficient vertical constraints, overly open environments, or facing complex three-dimensional structures, and many researchers have been working on it.

For z-axis drift situation, subsequent work such as LeGO-LOAM\cite{shan2018lego} significantly reduces vertical errors by decoupling ground points for pitch and roll angle optimization. LIO-SAM\cite{shan2020lio} further improves z-axis stability through IMU pre-integration factor fusion and gravity-alignment constraints. Recent research like BALM\cite{liu2021balm} proposes a Bayesian inference-based global optimization framework achieving high level vertical accuracy in GPS-denied indoor environments. Additionally, LIPS\cite{wang2022lips} addresses vertical drift in low-texture environments by enhancing ground segmentation with point cloud intensity information.

Many systems fail in feature-sparse or open environments due to feature matching challenges. V-LOAM\cite{zhang2017vloam} proposes a tightly-coupled visual-LiDAR approach that supplements sparse LiDAR data with visual features. Recent work like R3LIVE\cite{lin2022r3live} constructs radiance field maps to enable robust localization in unstructured environments using reflectance consistency constraints. Furthermore, SC-LeGO-LOAM\cite{shan2018lego, liu2022improved} integrates Scan Context descriptors to improve loop closure detection success rate by 40\% in repetitive structural scenarios like parking garages.

Multi-floor and complex 3D structures pose also significantly challenges for LiDAR SLAM. Ground-SLAM\cite{kim2020groundslam} exploits grounds in structured multi-floor environments to compress the pose drift mainly caused by LiDAR measurement bias. PALoc\cite{chen2021paloc} employs point cloud elevation distribution histograms for floor change recognition, combined with multi-floor topological mapping for cross-level localization. To address complex structures, LVI-SAM\cite{shan2021lvisam} fuses LiDAR-visual-inertial data to resolve sensor degradation in helical structures through multi-sensor collaboration. Additionally, MULLS\cite{pan2021mulls} proposes multi-scale local linear embedding descriptors, achieving 97\% localization success rate in complex scenarios like overpass structures.

\subsection{Perception Aware Trajectory and Yaw Planning}
Perception-aware trajectory and yaw planning has emerged as a critical research direction, aiming to harmonize motion planning with sensor's perceptual constraints to better adapt to environmental uncertainty and enhance navigation safety. Zhou \textit{et al.} \cite{RAPTOR} propose a planning framework which integrates perception-aware planning with risk-aware trajectory refinement by optimizing yaw angles to maximize map information gain. In \cite{Costante2017}, a direct vision-based approach is proposed to prioritize texture-rich regions and minimize localization uncertainty. Tordesillas \textit{et al.} \cite{PANTHER} define a dynamical visibility that adjusts trajectories to keep moving obstacles within the sensor FOV while avoiding collisions, demonstrating real-time performance in cluttered scenarios. Takemura \textit{et al.} \cite{takemura2022perception} denote perception quality as the density of the feature points, which is benefit with pose estimation. In the planning stage, the framework generates lots of candidate paths by rapidly-exploring random trees algorithm and selects one from them. Their another work \cite{takemura2023energy} proposed a framework consists of the global and local path planning respectively, which considering the energy consumption based on the aerodynamic model and the perception quality based on the scanning effect of the feature points around the UAVs. 

While existing approaches have made progress in specific scenarios, two critical challenges remain unresolved: (1) persistent vertical drift in open-top environments lacking ceiling features with single LiDAR, and (2) yaw optimization under directional perception constraints in trajectory planning with LiDAR. These limitations motivate our design of EAROL's tilted LiDAR configuration and coupled trajectory-yaw optimization framework.

\section{System Overview}

The overall system architecture is shown in Fig. \ref{zongtu}. We proposed a novel \textbf{Platform} with downward-mounted and tilted LiDAR installation, as shown in Fig. \ref{zongtu}(A). This strategic placement empowers the drone to simultaneously acquire lower-section geometric features for altitude estimation and perceive frontal environmental data for navigation. The airframe architecture adopts a modular three-tier design that optimizes both functionality and maintainability. More detailed information of hardware component and parameter will be illustrated in the experiment section.
Regarding the software architecture, as shown in Fig. \ref{zongtu}(B)-(D), our system features two core innovations: \textbf{Perception} and \textbf{Planning}. We develop an improved LIO system leveraging iterative error-state Kalman filter (IESKF) formulation with dynamic covariance adaptation. Furthermore, we develop a hierarchical planning framework that systematically integrates trajectory optimization with dynamic feasibility, smoothness, energy efficiency and collision-avoidance, while simultaneously generating optimal yaw angle sequences through a novel multi-objective optimization strategy that achieves effective balance between environmental exploration and target tracking missions.



\section{LiDAR-Inertial Odometry}
In this section, we proposed a LIO system based on IESKF. By means of a tightly-coupled sensor fusion framework, combined with the adaptive LiDAR tilt angles and dynamic motion compensation, we can fully exploit the advantages of the inverted and tilted LiDAR. 

\subsection{Adaptive LiDAR Tilt Angle Compensation}
To meet the requirements of perception and ensure the detection of most targets in front, a LiDAR tilt angle of 20° is selected. However, during the installation process, machining errors are inevitable and can lead to variations in the LiDAR tilt angle. If left unaddressed, this could affect localization accuracy. To enable the LiDAR system to perform adaptive initialization and automatically detect the tilt angle upon each startup, we proposed a method which utilizes the acceleration data measured by the IMU to calculate the angle at which the LiDAR is installed, and employs filtering algorithms to smooth and denoise the acceleration data, thereby enhancing the precision and robustness of the angle calculation.

Assuming that in a stationary state, the acceleration along the x-axis, y-axis, and z-axis are denoted as $a_x$, $a_y$ and $a_z$ respectively. Considering the direction of the gravitational acceleration, the tilt angle of the device can be estimated based on the ratio of the acceleration components in the XoZ plane as
\begin{equation}
pitch = \arctan{\frac{a_x^{filtered}}{a_z^{filtered}}}
\end{equation}
where $a_x^{filtered}$ and $a_z^{filtered}$ represent the low-pass filtered acceleration data from the IMU, which is applied to counteract the noise or irregular fluctuations that may be present in the raw IMU acceleration measurements, denoted as
\begin{equation}
\begin{aligned}
a_x^{filtered}(k) = \alpha a_x(k) + (1-\alpha) a_x^{filtered}(k-1) \\
a_z^{filtered}(k) = \alpha a_z(k) + (1-\alpha) a_z^{filtered}(k-1)  
\end{aligned}
\end{equation}

\subsection{LIO based on IESKF}
This section presents a LIO system based on IESKF, which significantly enhances the localization accuracy and robustness of UAVs in high-speed motion and complex environments through a tightly coupled sensor fusion framework, combined with dynamic motion compensation and adaptive handling of degenerate scenarios. There are three core ideas. Firstly, To address the discrepancies in IMU data and point cloud distortions caused by the rapid flight of UAVs, a Bézier curve interpolation method is proposed to generate smooth motion trajectories, and high-precision point cloud denoising is achieved through timestamp alignment strategies. Secondly, A degeneration detection method based on constrained normal vectors is introduced, which combines entropy quantification of environmental structure constraints to dynamically adjust map resolution through adaptive voxel filtering, suppressing localization drift caused by degeneration. Last but not least, IESKF is employed to fuse IMU pre-integration and LiDAR observations for state optimization. A multi-level voxel hash table with dynamic activation mechanism is utilized for efficient management of the global map.

\subsubsection{State Estimation and Dynamic Compensation}
State vector of the system is denoted as 
$$
\xi = [\mathbf{t}_W^I, \phi_W^I, \mathbf{v}_W^I, \beta_a, \beta_g, \gamma_W]^\top
$$
where $\mathbf{t}_W^T \in \mathbb{R}^3$ represents the position of IMU in the world coordinate, $\phi_W^I \in SO(3)$ represents the attitude of IMU (in the form of Lie algebra), $\mathbf{v}_W^I \in \mathbb{R}^3$ represents the velocity of IMU, $\beta_a, \beta_g \in \mathbb{R}^3$ represent the bias of accelerometer and gyroscope, $\gamma_W \in \mathbb{R}^3$ represents the correction term of gravity. Decompose the raw data of IMU into ground truth and noise, denoted as
$$
\begin{aligned}
&\tilde{\omega} = \omega_{\text{true}} + \beta_g + \epsilon_g, \quad \epsilon_g \sim \mathcal{N}(0, \Sigma_g)\\
&\tilde{\mathbf{a}} = \mathbf{R}_I^W (\mathbf{a}_{\text{true}}-\gamma_W)+\beta_a+\epsilon_a, \quad \epsilon_a \sim \mathcal{N}(0, \Sigma_a)
\end{aligned}
$$
The state is propagated through the pre-integration stage as 
\begin{equation}
\begin{aligned}
&\Delta \phi = \int_{t_k}^{t_{k+1}}(\tilde{\omega}-\beta_g) dt \\
&\Delta \mathbf{t} = \int_{t_k}^{t_{k+1}}\mathbf{v}_W^Idt + \frac{1}{2}\int_{t_k}^{t_{k+1}}\mathbf{R}_I^W(\tilde{\mathbf{a}}-\beta_a)dt^2\\
&\Delta \mathbf{v} = \int_{t_k}^{t_{k+1}}\mathbf{R}_I^W(\tilde{\mathbf{a}}-\beta_1)dt
\end{aligned}
\end{equation}
and the global update stage as
\begin{equation}
\begin{aligned}
&\mathbf{t}_W^{I_{k+1}} = \mathbf{t}_W^{I_k} + \mathbf{v}_W^{I_k}\Delta \mathbf{t} + \Delta \mathbf{t} \\
&\phi_W^{I_{k+1}} = \phi_W^{I_k} \oplus \Delta\phi\\
&\mathbf{v}_W^{I_{k+1}} = \mathbf{v}_W^{I_k}+\Delta \mathbf{v}
\end{aligned}
\end{equation}
where $\oplus$ represents the addition operation in Lie algebra, ensuring the manifold property of attitude updates.

To eliminate point cloud distortion caused by the abrupt movements, we propose an interpolation method based on Bézier curves. For a laser point with a timestamp of $\tau_j \in [t_k, t_{k+1}]$, we define the normalized time offset as 
$$
\mu = \frac{\tau_j-t_k}{t_{k+1}-t_k}
$$
The interpolation angular velocity and acceleration is defined as
\begin{equation}
\begin{aligned}
&\omega_{\text{bezier}} = (1-\mu)^2\omega_k+2\mu(1-\mu)\omega_m+\mu^2\omega_{k+1}\\
&\mathbf{a}_{\text{bezier}} = (1-\mu)^2\mathbf{a}_k+2\mu(1-\mu)\mathbf{a}_m+\mu^2\mathbf{a}_{k+1}
\end{aligned}
\end{equation}
where $\omega_m, \mathbf{a}_m$ is the data of IMU in the middle timestamp. By utilizing the interpolated motion parameters, the points in the LiDAR coordinate are transformed into the world coordinate as
\begin{equation}
\mathbf{p}_{\text{undistort}}^j = \underbrace{\text{exp}(\phi_W^I(\tau_j))}_{\text{rotation}} (\mathbf{p}_L^j+\mathbf{t}_L^I)+\mathbf{t}_W^I(\tau_j)+\mathbf{v}_W^I(\tau_j)\Delta\tau_j
\end{equation}
which can effectively eliminate the point cloud stretching or compression caused by movement.
\subsubsection{Residual Construction and Degradation Scenario Handling}
In order to the robustness of matching, the Mahalanobis distance weighting is introduced when constructing point-to-plane residuals as
\begin{equation}
r_j = \frac{\mathbf{n}_j^\top(\mathbf{p}_{\text{undistort}}^j-\textbf{c}_j)} {\sqrt{\mathbf{n}_j\top\Sigma_j\mathbf{n}_j}}
\end{equation}
where $\mathbf{n}_j$ is the normal vector of local plane, $\textbf{c}_j$ is the center point of plane, $\Sigma_j$ is the fitting covariance matrix of plane. By calculating the eigen value $\lambda_1\geq\lambda_2\geq\lambda_3$ of local point cloud covariance matrix via PCA, we can define the entropy as 
\begin{equation}
\textbf{H} = -\sum^3_{i=1}\gamma_i \ln \gamma_i, \quad \gamma_i = \frac{\lambda_i}{\Sigma\lambda_i}
\end{equation}
where $\textbf{H} < \textbf{H}_{th}$ means the scenario is degenerate.

To apply a damping term $\textbf{H} \leftarrow \textbf{H} + \lambda\mathbf{p}_{\text{degen}}$ to the degenerate directions during state updates, thereby suppressing the accumulation of errors in unobservable directions, we construct a projection matrix as
\begin{equation}
\mathbf{p}_{\text{degen}} = \sum_{i:\gamma_i<\theta_{th}} \mathbf{v}_i\mathbf{v}_i^\top
\end{equation}

We adjust the voxel resolution based on the current point cloud density $N_\text{current}$ and the target density $N_{\text{target}}$ to perform dynamic voxel filtering, reducing mismatches by lowering the resolution in degenerate areas as
\begin{equation}
d_{\text{new}}=d_{\text{old}}\cdot\text{exp}(-\eta\frac{N_\text{current}-N_{\text{target}}}{N_{\text{target}}})
\end{equation}

\subsubsection{IESKF Optimization and Map Update}
We define the error state vector as 
$$
\delta\xi=[\delta\mathbf{t}_W^I, \delta\phi_W^I, \delta \mathbf{v}_W^I, \delta\beta_a, \delta\beta_g, \delta\gamma_W]^\top
$$
and then we linearize the observation model by Jocobian matrix $\textbf{J}_h$ to update state as
\begin{equation}
\delta\xi^{(i+1)} = \delta\xi^{(i)}-(\textbf{J}_h^\top\mathbf{R}^{-1}\textbf{J}_h+\mathbf{p}^{-1})^{-1}\textbf{J}_h^\top\mathbf{R}^{-1}r^{(i)}
\end{equation}
After updating the state, we mapping it to manifold space by $\xi \leftarrow \xi \oplus \delta \xi^{(i+1)}$ to ensure the stability of values.

In the mapping, we employ a dual-layer voxel hash table with both fine and coarse resolutions to store point cloud data. The fine layer is dedicated to preserving high-frequency details, facilitating precise local matching. Conversely, the coarse layer is responsible for maintaining the global structure, enabling swift data retrieval. Furthermore, we implement a dynamic activation mechanism that selectively updates only the voxels in the vicinity of the UAV, thereby significantly reducing computational overhead.

\subsubsection{Relocalization Module}
By encoding local environmental features using spherical harmonic functions, after matching candidate regions, we combine pose graph optimization to achieve rapid relocalization as
\begin{equation}
f(\mathbf{p})=\sum_{l=0}^{L}\sum_{m=-l}^{l}a_{lm}Y_l^m(\theta,\phi)
\end{equation}
where $a_{lm}$ is the factor, $\theta$ is the angle between $\mathbf{p}$ and z-axis, $\phi$ is the angle between $\mathbf{p}$ and XoY plane.

Thus, we can adopt robust and accurate localization for other modules.

\section{Trajectory and Yaw Angle Generation}
In this section, a hierarchical optimization approach is proposed for trajectory and yaw angle generation, which consider collision-avoidance and control effort of the trajectory while optimizing the yaw angle to achieve balance between exploring the environment and tracking the target.

\subsection{Trajectory Generation}

For the UAV trajectory generation, the overall objective is to generate a smooth trajectory $\mathbf{p}(t)\colon\left[0,t_M\right]\rightarrow \mathbb{R}^3$, which minimize the control efforts represented by the $s$-th order derivative $||\mathbf{p}^{(s)}||_2^2$, while also satisfying all dynamic constraints and obstacle-avoidance constraints. Considering the differential flatness characteristic, multistage polynomial curves can be used to parameterize the trajectory, as proposed in \cite{GCOPTER}. The trajectory $\mathbf{p}(t)$ is composed by $\mathbf{M}$ pieces of polynomial curves with a degree of $N=2s-1$. When giving the initial state $\textit{state}_0 = \mathbf{p}^{(0:s-1)}(0)$ and the final state $\textit{state}_f = \mathbf{p}^{(0:s-1)}(t_M)$, the $i$ piece of curve is parameterized by inner waypoints $\mathbf{Q}=(\mathbf{q}_1,...,\mathbf{q}_{M-1})$ and time allocation $\mathbf{T}=(T_1,...,T_{M-1})$ as follows:

\begin{equation}
\mathbf{p}_i(t)=\mathbf{C}_i(\mathbf{Q},\mathbf{T},\textit{state}_0,\textit{state}_f)\beta(t)
\end{equation}
where $\beta(t) = [1,t,t^2,...,t^N]^\top \in \mathbb{R}^{2s}$ is the basis function of the polynomial curve and $\mathbf{C}_i=[c_{i0},c_{i1},...,c_{iN}]^\top \in \mathbb{R}^{2s}$ is the coefficient matrix of the curves. Specifically,  $\mathbf{C}_i$ can directly obtain by solving a linear system from the known $\mathbf{Q}$ and $\mathbf{T}$, according to \cite{GCOPTER}.

The trajectory generation problem can be formulated as a nonlinear optimization problem as follows:

\begin{equation}
\begin{split}
&\min_{Q,T} \sum_i \lambda_iJ_i\\
\end{split}
\end{equation}
where $J_i$ are several penalty terms and $\lambda_i$ are corresponding weighs, consisted of trajectory smoothness penalty $J_s$, dynamic feasibility penalty $J_f$, obstacle collision penalty $J_o$ and overall time penalty $J_t$.

\subsubsection{Smoothness penalty $J_s$} 
As illustrated above, the smoothness penalty is the integral of position's $s$-th order derivative:

\begin{equation}
J_s=\sum_{t_0}^{t_M} \|{\mathbf{p}^{(s)}(t)}\|_2^2
\end{equation}

\subsubsection{Dynamic feasibility penalty $J_f$}
We limit the velocity, acceleration, and jerk when exceed the given physical thresholds:
\begin{equation}
J_f= J_v+J_a+J_j
\end{equation}
\begin{equation}
J_v =\sum_{t_0}^{t_M} \max((\dot{\mathbf{p}}(t_i)^2-vel_{max}^2),0)
\end{equation}
\begin{equation}
J_a =\sum_{t_0}^{t_M} \max((\ddot{\mathbf{p}}(t_i)^2-acc_{max}^2),0)
\end{equation}
\begin{equation}
J_j =\sum_{t_0}^{t_M} \max((\dddot{\mathbf{p}}(t_i)^2-jer_{max}^2),0)
\end{equation}
where $vel_{max}$, $acc_{max}$, $jer_{max}$ denote the maximum limit of the velocity, acceleration, and jerk corresponding to the actuator.

\subsubsection{Obstacle collision penalty $J_o$}
When the distance to the nearest obstacle is too small, it means a higher probability of having a collision. Therefore, we conduct penalty when the closest distance is less than a given threshold:
\begin{equation}
J_o =\sum_{t_0}^{t_M} \max(\mathcal{D}-dis(\mathbf{p}_i),0)
\end{equation}
where $dis(\cdot)$ denotes the distance from the current position to the nearest obstacle and $\mathcal{D} >$ 0 is the threshold decided when to penalize.

\subsubsection{Overall time penalty $J_t$}
We hope the total flight time can be as short as possible, although within the constraints of feasibility and smoothness:

\begin{equation}
J_t =\sum_{t_0}^{t_M} T_i
\end{equation}

The nonlinear optimization problem above can be solved efficiently by unconstrained optimization algorithms such as L-BFGS. 

\subsection{Yaw Angle Optimization}

\begin{figure}[t]
\centering
\includegraphics[width=0.485\textwidth]{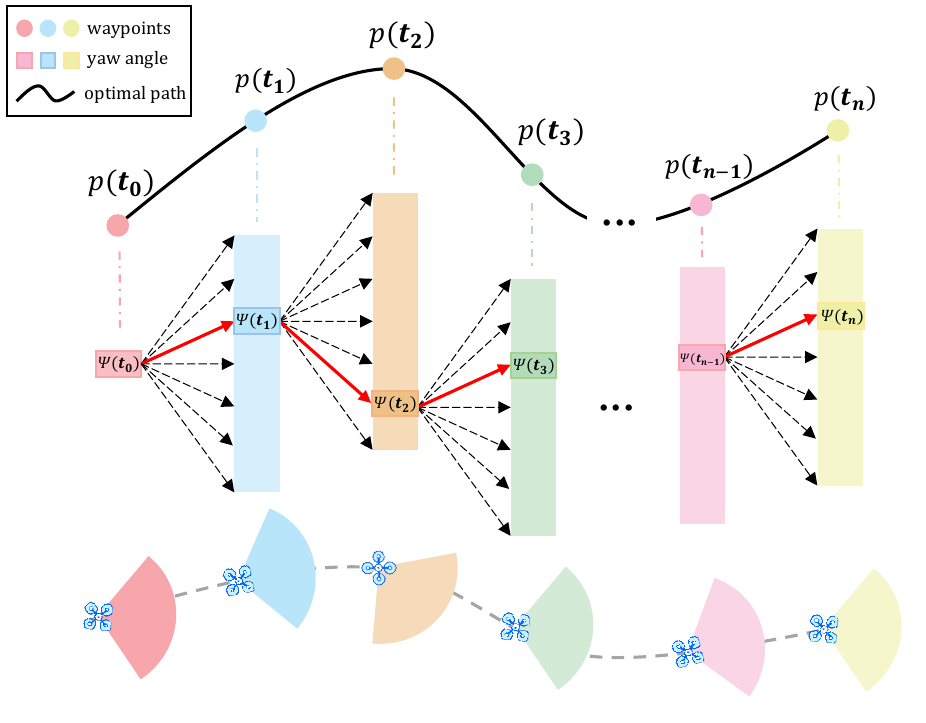}
\caption{The graph for searching yaw sequence. We divide the trajectory within the horizon into n segments. Based on the yaw angle found in the previous search layer, we sample within a certain range to generate the search space, as represented by the wider stripes in the figure.}       
\label{graph search}       
\end{figure}

For the generation of the yaw angle sequence, we sample target points along the trajectory at equal time intervals. Each target point is sampled by yaw angles from $-\pi$ to $\pi$ at the same interval to generate various nodes, which can be represented as $(p_x, p_y, p_z, \psi, t)$. The resulting graph, as shown in Fig. \ref{graph search}, features nodes within the same layer sharing the same timestamp and three-dimensional position, with yaw angles uniformly distributed between $-\pi$ and $\pi$. To fully consider the scene requirements, a cost function is constructed as
\begin{equation}
C = \alpha \cdot \sum H_i - \beta \cdot T(\psi) - \gamma \cdot E(\psi)
\label{cost function}
\end{equation}
where $\alpha$, $\beta$ and $\gamma$ are dynamic coefficients, which satisfy $\alpha+\beta+\gamma=1$, utilized to adjust the weightings of various components. Equation (\ref{cost function}) measures the degree of environmental exploration using the information entropy of a grid map, assess the recognition and tracking effectiveness by the polar angle of the target within the FOV, and add a yaw rotation cost term to balance energy consumption.

The first term is about environmental exploration. Assume that the occupancy probability of each grid is denoted as $p_i \in [0,1]$, the information entropy is denoted as
\begin{equation}
H_i = e^{-R_i} [-p_i \log_2 p_i-(1-p_i) \log_2 (1-p_i)] 
\label{entropy formula}
\end{equation}
where $-e^{R_i}$ represents the negative exponential term of the distance, used to attenuate the influence of distant grids. We employ a local map centered on the UAV with a side length of 5 meters, and statistically analyze the information entropy within the sectoral region based on the polar coordinate angle centered on the UAV.

The second term is about recognition and tracking error, which is measured by the square of the polar coordinate angle deviation between the target position and the center of the FOV, to assess the target following performance. This value is normalized and mapped to the interval $[0,1]$, as
\begin{equation}
T(\psi) = (\frac{\psi(dx,dy)}{\pi})^2
\label{tracking formula}
\end{equation}
where $dy$ and $dx$ are the offset of the target relative to the center of the FOV. To prevent frequent fine-tuning of the yaw angle, a dead-zone threshold $\psi_{th}$ is set as
$$
\psi(dx,dy) = 
\begin{cases}
0,\quad & |\arctan(\frac{dy}{dx})| \leq \psi_{th} \\
\arctan(\frac{dy}{dx}),\quad & |\arctan(\frac{dy}{dx})| > \psi_{th}
\end{cases} 
$$

The third term is about balancing the energy consumption of yaw rotation, represented as the square of the difference from the previous moment
\begin{equation}
E(\psi) = (\psi_t - \psi_{t-1})^2
\end{equation}

The graph search problem can be solved using Dijkstra algorithm. To address the large scale of the graph and the redundancy of information, we employ a pruning strategy based on the maximum angular velocity. By utilizing the current yaw angle, the maximum angular velocity, and the sampling interval of the trajectory, we can calculate the feasible range of the yaw angle at the next moment under dynamic constraints. Sampling only within this range significantly reduces the size of the graph. Furthermore, considering the collision avoiding trajectory is constantly evolving, we adopt a horizon for incremental yaw sequence search.

Thus, we can adopt collision-avoidance trajectory binding with search-tracking balanced yaw angle sequence.

\section{EXPERIMENTS}

We performed three experiments in physical environments to validate the odometry and planning module of EAROL. Experiments include mapping in an indoor maze without ceilings, mapping in an outdoor large scale scenario on the playground with several challenging elements, as well as tracking dynamic object in indoor challenging environment.

\subsection{Platform}

The drone platform, as shown in Fig. \ref{zongtu} (A), has a 350mm wheelbase and 8.5-inch carbon fiber propellers. The onboard computer is an Intel NUC13ANKi7, equipped with an Intel Core i7-1360P processor, 16GB RAM, and 512GB ROM. The flight control unit is Holybro Pixhawk 6C, featuring an STM32H743, 32-bit Arm Cortex-M7, 480MHz, 2MB memory, and 1MB SRAM. The LiDAR is MID360, whose FOV is -7° to 52°, obtaining more than 200,000 points per second. The dynamic obstacle consists of a ring and an unmanned ground vehicle (UGV), where the ring can be easily recognized using RANSAC (RANdom SAmple Consensus), and the UGV with max velocity 1.5m/s. The specific physical data of the platform are shown in Table \ref{physical data}.
\begin{table}[t] 
\setlength{\tabcolsep}{10mm}{
        \caption{Physical Data of EAROL Platform}
        \label{physical data}
        \begin{center}
        \begin{tabular}{c  c}
        \toprule[1.2pt]
        \textbf{Param.} & \textbf{Data}\\
        \hline
        Mass [kg] & 1.89\\
        
        Maximum Thrust [N] & 50.485\\
        
        Thrust-to-Weight Ratio & 2.67\\
        
        Inertia [g·m$^2$] & [3.69, 3.02, 5.54]\\
        
        Battery Endurance [min] & $>$15\\
        \bottomrule[1.2pt]
        \end{tabular}
        \end{center}
        }
\end{table}

\subsection{Indoor Localization}

\begin{figure}[t]
\centering
\includegraphics[width=0.485\textwidth]{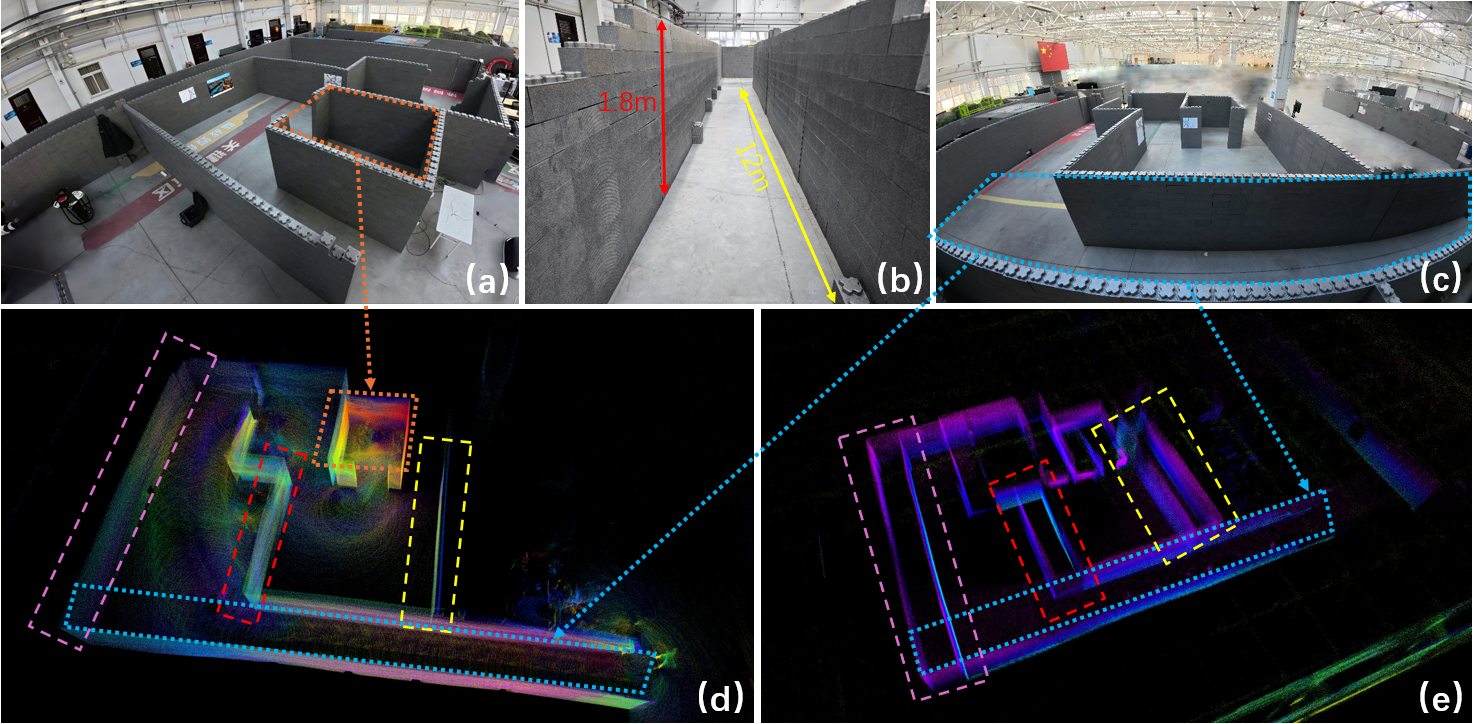}
\caption{Scenarios and results of indoor localization experiment. (a) and (c) present comprehensive views of our custom-built maze structure, while (b) details the constructed narrow corridor. (d) and (e) present the mapping results of EAROL and FAST-LIO2 respectively, where corresponding area across are highlighted using “- -” type identical color bounding boxes. The same area in real-world photos and point cloud maps are using “$\cdot\cdot$” type.}       
\label{exp1}       
\end{figure}

To demonstrate EAROL's robustness in degenerate environment, we constructed a challenging small-scale scenario spanning 17 meters in length and 12 meters in width in the first experimental setup. This environment incorporates a 12-meter-long narrow corridor and point clouds above 1.5 meters are filtered  to simulate extreme conditions for LIO system. For comparative analysis, both our EAROL and a conventional LIO framework (utilizing FAST-LIO2 \cite{xu2022fast}) are installed on identical mobile platforms. All the hardware configurations are the same, except the mount of LiDAR. The mapping results displayed in Fig. \ref{exp1} demonstrated the performance contrast between EAROL and FAST-LIO2 under identical experimental conditions. Visual analysis as shown in Fig. \ref{exp1} (d) and (e) with different color frames, reveal that our algorithm produced exceptionally clear and precise mapping outputs with well-defined boundaries, accurately reconstructing the spatial configuration of the environment. In contrast, the FAST-LIO2 implementation exhibited significant positional drift and substantial deviations in boundary representation, failing to maintain structural fidelity to the actual testing environment.

\begin{figure}[t]
\centering
\includegraphics[width=0.485\textwidth]{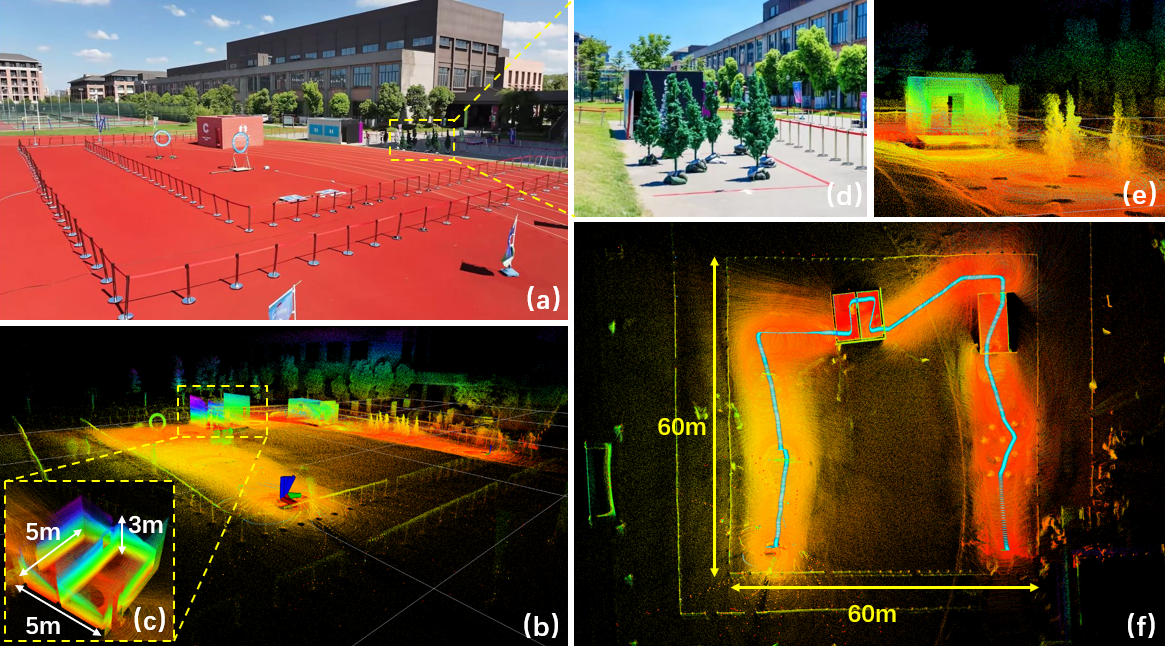}
\caption{Scenarios and results of outdoor localization experiment. (a) and (d) present the panoramic view and dense vegetation clusters of the scenario respectively, while (b),(c) and (d) display panoramic, maze and dense vegetation clusters mapping results. (f) illustrates both the aerial mapping outcome from a top-down perspective and the associated navigation trajectory.}       
\label{exp2}       
\end{figure}
\begin{figure}[t]
\centering
\includegraphics[width=0.485\textwidth]{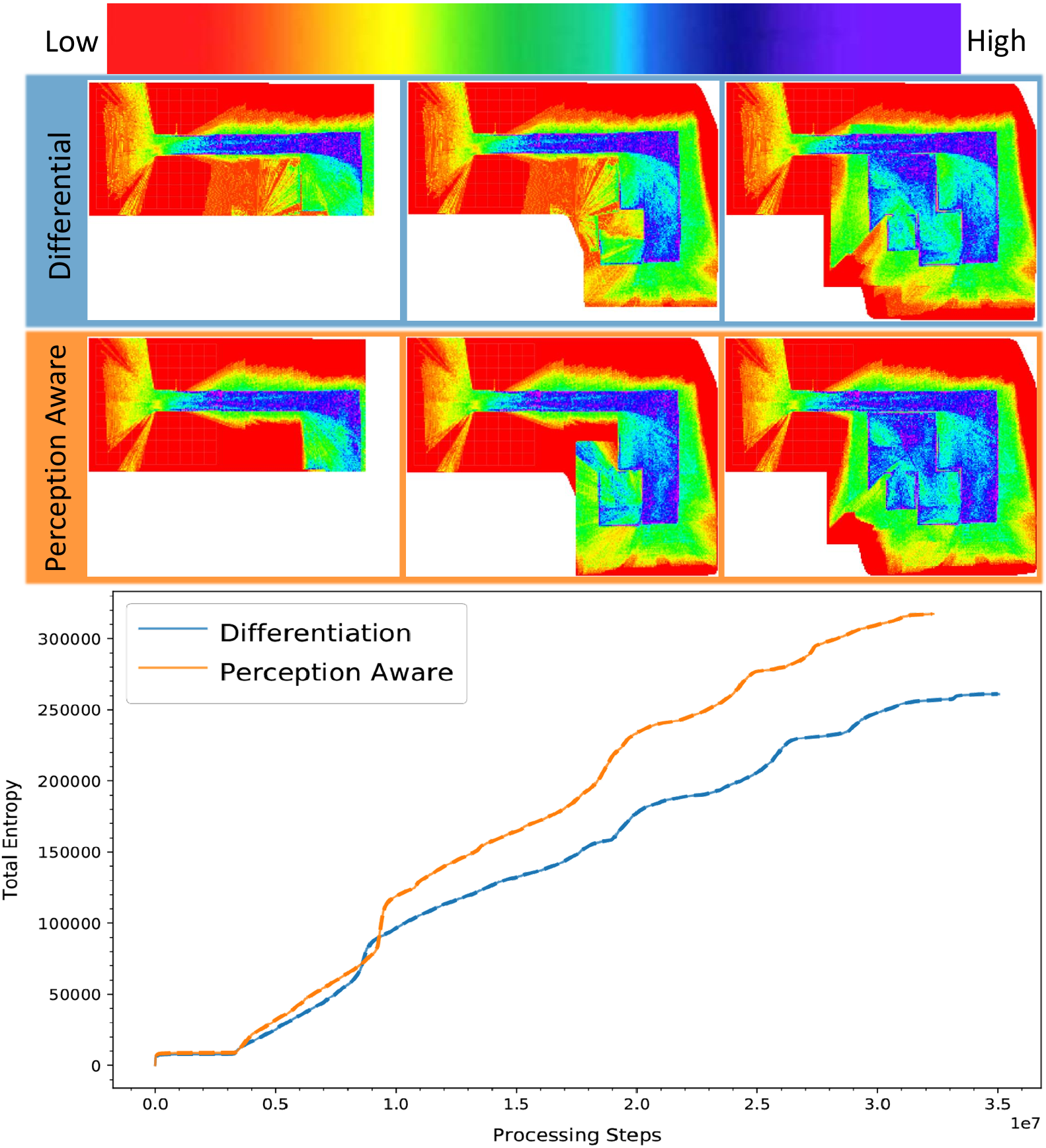}
\caption{Information entropy of the scenario over task steps. The 6 figures on the top are entropy maps at different steps. The entropy value of each pixel is derived by adding the entropy values of grids at the same horizontal coordinates but different heights in the three-dimensional map. Red represents low entropy, and purple represents high entropy.}       
\label{entropy}       
\end{figure}
\subsection{Outdoor Localization}

To rigorously validate scalability of EAROL, we conducted large-scale mapping experiments within a scenario spanning 60 meters in length and width. The test environment incorporates five distinct challenge elements: (1) dense vegetation clusters, (2) enclosed shipping containers, (3) an open-air maze structure with 3-meter-high walls, (4) stationary rings, and (5) dynamically moving rings. During autonomous navigation, EAROL demonstrated real-time environmental perception through synchronized data acquisition and adaptive processing, enabling high-fidelity 3D reconstruction while maintaining precise obstacle negotiation capabilities. Fig. \ref{exp2} comprehensively documents the mission execution process, demonstrating remarkable congruence between the aerial trajectories and reconstructed spatial features with ground truth environmental configurations. Indoor localization and outdoor localization achieved robust performance in open-top environments, particularly in maintaining high level positioning accuracy and mapping.

\subsection{Tracking Dynamic Object in the Maze}

\begin{figure}[t]
\centering
\includegraphics[width=0.485\textwidth]{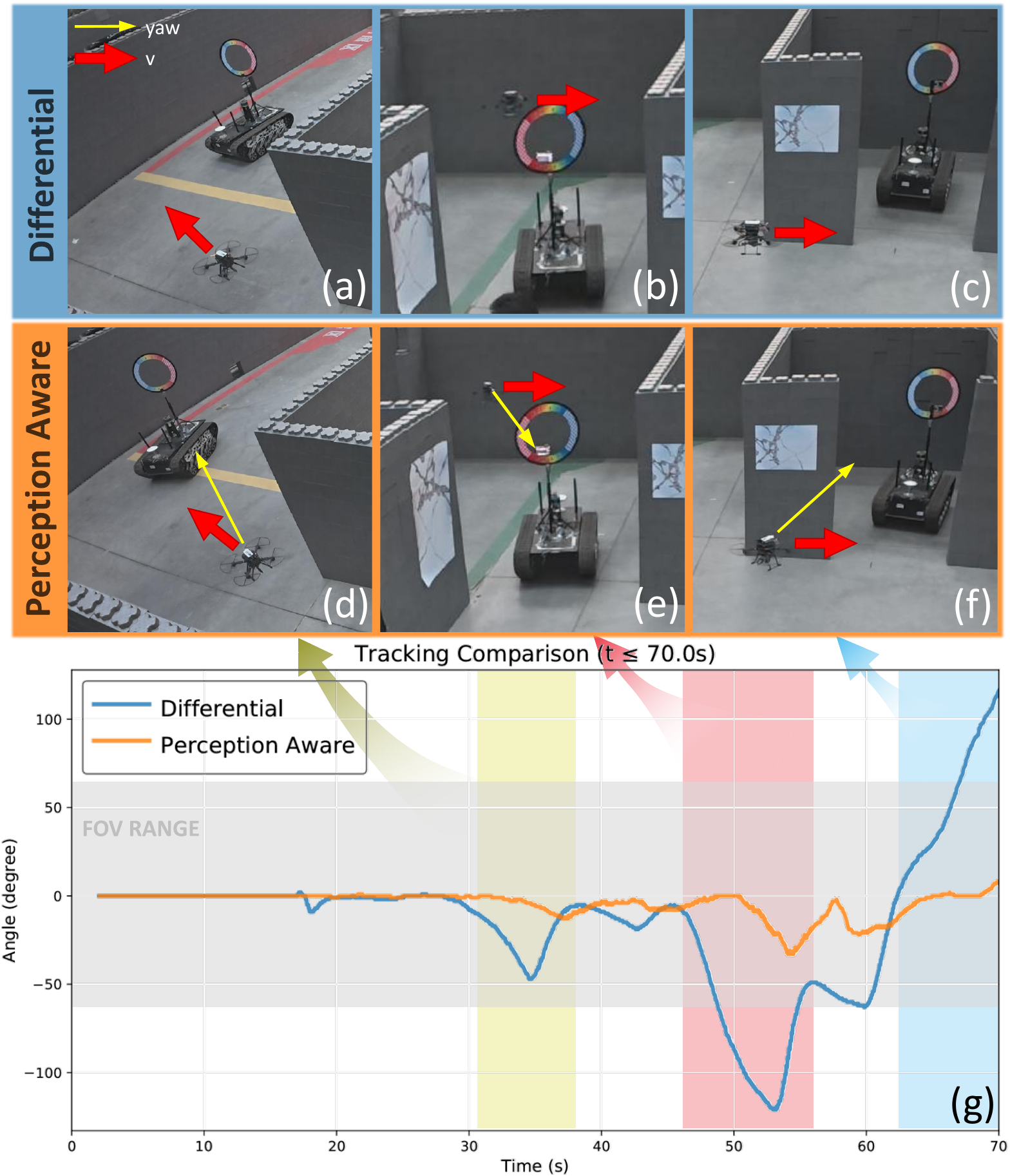}
\caption{Tracking error over times. Figures (a)-(c) and (d)-(f) depict the image captured during the execution of the yaw generated by differential and perception aware respectively, where red and yellow arrows denote the velocity direction and orientation of the yaw respectively (in the case of differential generation, the velocity direction and yaw angle orientation are consistent). In figure (g), the yellow, red, and blue regions correspond to the real-world at various instances in time, while the gray range represents the FOV.}
\label{tracking}       
\end{figure}

We conducted dynamic target tracking experiments within the indoor scenario mentioned in the first experiment using the proposed LIO, in which parameters are set as $\alpha=0.3, \beta=0.6, \gamma=0.1$ in (\ref{cost function}). The platform initiated its flight from one end of the tunnel and is required to acquire and track a dynamic target (comprising a ring mounted on an UGV) during its navigation. After acquiring the target, EAROL generated yaw sequence based on entropy map, tracking error and energy consumption for keeping the target in the FOV as well as avoiding collision. A comparative analysis was conducted between differential-based yaw generation and EAROL, with quantitative evaluation performed through two critical metrics. Global map information entropy evolution is updated by (\ref{entropy formula}) and LIO, using local grid map updated by raycast. Tracking error distribution defined as (\ref{tracking formula}). Fig. \ref{entropy} demonstrated 6 figures of entropy maps at different steps, and finally a 22$\%$ enhancement in environmental perception quality. Fig. \ref{tracking} demonstrated the drone’s ability to effectively keep the target centered in its FOV (no more than 32 degrees) while executing perception-aware yaw angle adjustments, all without deviating from its intended path. In contrast, when utilizing differential generation for the yaw angle, the drone often loses sight of the target (more than 120 degrees), which is a consequence of the discrepancy between its travel direction and the target’s facing direction. Statistics showed that tracking error reduction rates of 81$\%$ in cumulative tracking error and 73$\%$ in maximum instantaneous error.

\section{CONCLUSIONS}
This work presents a comprehensive and environmental augmented solution to UAV localization and planning in open-top degenerate scenes. By innovatively downwards mounting and tilting the LiDAR, EAROL eliminates vertical drift through ground geometric constraints while maintaining forward perception capabilities. The proposed LIO system achieves high level accuracy in several extreme scenarios. Furthermore, our perception-aware yaw planning algorithm optimizes environmental exploration, target tracking, and energy efficiency through graph search with entropy-driven cost functions. Finally, we conducted physical experiment to demonstrate the system's robustness and efficiency. This study offers a scalable solution for UAV applications in search-and-rescue and exploration missions. Future research will focus on optimizing flight velocity while integrating semantic-aware modules to establish a multi-modal sensing framework that enhances environmental perception under dynamic operational constraints.

\addtolength{\textheight}{-12cm}   





\bibliographystyle{IEEEtran}
\bibliography{Citation}

\end{document}